\documentclass[11pt]{article}

\usepackage[final]{acl}

\usepackage{times}
\usepackage{latexsym}

\usepackage[T1]{fontenc}

\usepackage[utf8]{inputenc}

\usepackage{microtype}

\usepackage{inconsolata}

\usepackage{graphicx}

\usepackage{booktabs}
\usepackage{multirow}
\usepackage{xcolor}
\usepackage{subcaption}
\usepackage{graphicx}
\usepackage{array}
\usepackage{tikz}
\usetikzlibrary{positioning,arrows.meta,shapes.geometric}
\usepackage{graphicx}      
\usepackage{amsmath,amssymb} 
\usepackage{multirow}    
\usepackage{makecell}     
\usepackage[table]{xcolor} 
\definecolor{gaingreen}{RGB}{218, 238, 213}
\definecolor{lossred}{RGB}{250, 222, 220}
\usepackage{threeparttable}

\usepackage{tabularx}

\usepackage{xcolor}
\usepackage{listings}

\lstdefinestyle{promptstyle}{
  basicstyle=\ttfamily\scriptsize,
  breaklines=true,
  breakatwhitespace=false,
  columns=fullflexible,
  keepspaces=true,
  frame=single,
  framerule=0.4pt,
  rulecolor=\color{gray!60},
  backgroundcolor=\color{gray!4},
  xleftmargin=2pt,
  xrightmargin=2pt,
  aboveskip=0.6em,
  belowskip=0.6em,
  showstringspaces=false
}

\title{Structured Frequency-Domain Evidence for \\ LLM-Based Time-Series Anomaly Detection}

\author{
  Jungwook Seo\textsuperscript{1},
  Sangwon Son\textsuperscript{1},
  Minjeong Kim\textsuperscript{1},
  Seungmin Han\textsuperscript{1},
  Seojin Yoo\textsuperscript{2},
  Sungyong Baik\textsuperscript{1,2,\textdagger} \\
  \textsuperscript{1}Department of Artificial Intelligence, Hanyang University \\
  \textsuperscript{2}Department of Data Science, Hanyang University \\
  Seoul, Republic of Korea \\
  \texttt{\{zungwooker, swson, mjkim0720, handsomemin, dbtjwls0821, dsybaik\}@hanyang.ac.kr}
}

\begin{document}
\maketitle

\begingroup
\renewcommand{\thefootnote}{\textdagger}
\footnotetext{Corresponding author.}
\endgroup

\begin{abstract}
Time-series anomalies can appear not only as pointwise deviations but also as changes in recurring temporal structure, such as shifted periodicity or localized oscillatory fluctuations.
However, existing LLM-based time-series anomaly detection methods mainly expose time-domain evidence through indexed values, plots, or de-seasonalized representations, leaving spectral structure implicit.
We propose an evidence-augmented zero-shot TSAD framework that preserves indexed de-seasonalized observations while adding compact frequency-domain evidence computed with the Fast Fourier Transform (FFT).
The evidence is constructed at two resolutions: global frequency-domain evidence summarizes sequence-level periodic context, while local frequency-domain evidence captures time-localized spectral departures.
Experiments on AnomLLM with InternVL2-LLaMA3-76B, Qwen2.5-VL-72B-Instruct, Gemini-2.5-Flash, and GPT-4o, together with evaluation on the TSB-AD-U subset, show that explicit frequency-domain evidence improves LLM-based TSAD baselines.
These results suggest that frequency-domain evidence can complement indexed and de-seasonalized time-domain inputs for zero-shot LLM-based TSAD.
\end{abstract}

\section{Introduction}

Beyond isolated spikes or pointwise deviations, anomalous intervals can appear as subtle changes in recurring temporal structure, such as altered cycles, shifted periodic components, or unstable local fluctuations.
These patterns are naturally frequency-relevant: detecting them may require evidence about how the periodic structure of a sequence changes over time.

This issue is critical in realistic time-series anomaly detection (TSAD)~\cite{garcia2020review, pang2020review}, where anomaly labels, anomaly categories, and dataset-specific priors are often unavailable before deployment~\cite{han2022adbench}.
A zero-shot setting therefore requires the detector to localize anomalous intervals directly from the input sequence, without relying on benchmark-specific taxonomies, historical examples, or auxiliary supervision.
This is stricter than training-free inference alone, since a method can avoid task-specific model training while still using external priors.

Recent studies have investigated large language models (LLMs) for TSAD, showing that LLMs can localize anomalous intervals without task-specific training from numerical sequences and, in some cases, visual context~\cite{sigllm, llmad, anomllm, llmtsad}.
Existing LLM-based methods expose time series through textualized numerical sequences, visualized plots, or transformed inputs such as de-seasonalized sequences~\cite{sigllm,llmad,tama}, while LLM-TSAD preserves temporal indices to improve interval-level localization~\cite{llmtsad}.
These designs help LLMs reason over temporal order and interval boundaries, but they do not explicitly summarize spectral properties such as dominant periods, spectral energy distribution, or time-localized frequency changes.

\begin{table*}[t!]
\centering
\setlength{\tabcolsep}{4.5pt}
\renewcommand{\arraystretch}{1.2}

\resizebox{\textwidth}{!}{%
\begin{tabular}{l|l|ccc|cc}
\toprule
\textbf{Probe} & \textbf{Metric}
& \textbf{Indexed Text}
& \textbf{Text + Plot}
& \textbf{De-seasonalized}
& \textbf{Global Freq.}
& \textbf{Local Freq.} \\
\midrule
Single sinusoid
& Top-1 Recovery $(10\%)$
& 24.5 & 23.5 & 14.5 & \textbf{95.5} & 52.0 \\

Multi sinusoid
& Top-$k$ Recovery $(10\%)$
& 24.7 & 18.1 & 34.6 & \textbf{91.9} & 43.0 \\

Local frequency change
& Detection Acc.
& 11.5 & 80.5 & 88.0 & 8.5 & \textbf{90.5} \\

\bottomrule
\end{tabular}%
}
\caption{
\textbf{Summary of frequency-reasoning probes.}
Top-1 and Top-$k$ Recovery use a 10\% relative-error tolerance.
Global Frequency and Local Frequency denote indexed text augmented with global and local frequency summaries, respectively.
}
\label{tab:motivation-probes}
\vspace{-0.4cm}
\end{table*}

\begin{figure*}[t!]
    \centering
    \includegraphics[width=0.55\textwidth]{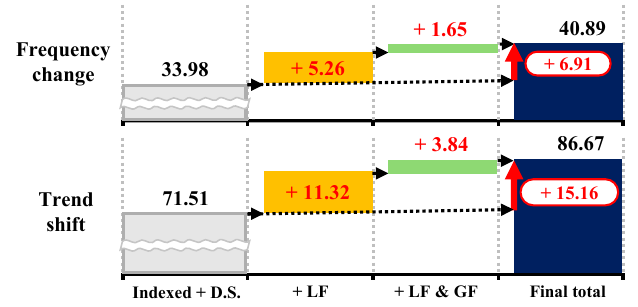}
        \vspace{-0.2cm}
    \caption{
\textbf{Stepwise F1 gains from local (LF) and global (GF) frequency-domain evidence on representative hard anomaly types.}
    The waterfall plot shows incremental gains over the indexed de-seasonalized baseline.
    }
    \label{fig:stepwise_gain}
    \vspace{-0.2cm}
\end{figure*}

Frequency-domain modeling has long been useful in TSAD for capturing periodic structure, long-range dependencies, and anomalies that are weakly expressed in local time-domain values~\cite{tfad, fselstm, fcvae, dualtf}.
However, existing frequency-aware TSAD methods typically encode spectral information inside model-specific architectures, reconstruction objectives, or supervised pipelines~\cite{tfad, fselstm}.
This creates a representation-level evidence gap: spectral information may be useful for anomaly localization, but it is rarely exposed as explicit input evidence that a general-purpose LLM can directly condition on.

We address this gap by constructing frequency-domain evidence at the input level.
The proposed framework augments indexed de-seasonalized observations with compact global and local frequency-domain evidence computed from the input sequence using the Fast Fourier Transform (FFT).
Global evidence captures the sequence-level periodic regime, while local evidence captures spectral behavior within overlapping temporal windows.
The resulting input representation allows the LLM to condition on indexed time-domain observations and explicit frequency-domain evidence without using predefined anomaly categories, auxiliary supervision, or task-specific model training.

\vspace{2pt}
\noindent\textbf{Our contributions are as follows:}
\begin{itemize}
\item We identify a representation-level evidence gap in LLM-based TSAD: existing inputs help represent temporal order and interval boundaries, but leave frequency-relevant cues largely implicit.
\item We propose a training-free evidence-augmented framework that constructs explicit global and local FFT-based summaries directly from the input sequence.
\item We validate the framework on the AnomLLM benchmark with multiple multimodal LLMs and the TSB-AD-U subset protocol used by LLM-TSAD, showing improvements over strong LLM-based TSAD baselines.
\end{itemize}

\section{Can LLMs Infer Frequency Structure from Time-Domain Inputs?}
\label{sec:motivation-frequency}

In time-series anomaly detection (TSAD), anomalous intervals may remain within a plausible value range while changing their period, frequency composition, or oscillatory regularity.
Such cases are difficult to identify from pointwise values alone because the anomalous pattern is expressed through changes in temporal structure rather than through large amplitude deviations.
This raises a representation-level question: can LLMs reliably recover frequency structure when it is provided only implicitly through time-domain inputs?

To examine this question, we conduct controlled probing experiments on synthetic sinusoidal sequences.
These probes are not intended as TSAD benchmarks, but as diagnostic tasks that isolate frequency reasoning from dataset-specific anomaly patterns.
Table~\ref{tab:motivation-probes} summarizes three probes: \emph{single sinusoid}, \emph{multi sinusoid}, and \emph{local frequency change}.
The first two probes evaluate sequence-level frequency recovery: the model must identify the dominant frequency in a single-component signal or recover the dominant components in a mixture of sinusoids.
The third probe evaluates localized frequency-change detection: the model must identify whether a segment contains a change in oscillatory behavior.

As shown in Table~\ref{tab:motivation-probes}, global frequency-domain evidence substantially improves sequence-level frequency recovery, achieving the best results on both the single- and multi-sinusoid probes. 
This result indicates that full-sequence spectral summaries provide periodic information that the LLM does not reliably recover from indexed values, plots, or de-seasonalized time-domain inputs alone. 
In particular, global summaries make the dominant periodic regime explicit, which is essential for reasoning about frequency components that persist across the sequence.

However, sequence-level frequency recovery alone is insufficient for anomaly localization. 
The same global evidence performs poorly on the local frequency-change probe because a full-sequence spectrum aggregates spectral content over time and therefore does not indicate where a frequency change occurs. 
Local frequency-domain evidence complements this limitation by exposing window-level changes in dominant frequency and spectral entropy, leading to the highest detection accuracy on the localized probe. 
Together, these results motivate a two-level evidence design: global summaries provide sequence-level periodic context, while local summaries expose time-localized spectral departures.

Figure~\ref{fig:stepwise_gain} shows that this diagnostic pattern also appears in TSAD performance. 
Starting from the indexed de-seasonalized baseline, local frequency-domain evidence improves frequency-relevant anomaly types by exposing window-level spectral changes, while global evidence provides additional sequence-level context. 
This connection between the probes and TSAD results supports the use of both evidence levels in the final framework.

\section{Preliminaries}
\label{sec:prelim}

\subsection{LLM-based TSAD}
We formulate time-series anomaly detection (TSAD) with a language model as an interval prediction task.
Let $\mathbf{x}=(x_1,\ldots,x_T)$ denote a univariate time series of length $T$, where $x_t$ is the observation at time index $t$.
The goal is to predict a set of anomalous intervals
$\hat{\mathcal{A}}=\{(\hat{s}_m,\hat{e}_m)\}_{m=1}^{M}$,
where $\hat{s}_m$ and $\hat{e}_m$ are the predicted start and end indices of the $m$-th anomalous interval, and $M$ is the number of predicted intervals.

Following LLM-TSAD~\cite{llmtsad}, the input sequence is provided as indexed text, where each observation $x_t$ is paired with its time index $t$.
When visual input is enabled, a time-series plot is additionally provided as an auxiliary modality.
The plot is not used as a replacement for indexed text, but as an additional representation in multimodal prompting.

\subsection{De-seasonalization-Based Prompting}
LLM-TSAD~\cite{llmtsad} combines index-aware prompting with de-seasonalization.
Given an additive decomposition
\begin{equation}
    x_t = s_t + \tau_t + r_t,
\end{equation}
where $s_t$ denotes the seasonal component, $\tau_t$ the trend component, and $r_t$ the residual component at time $t$, de-seasonalization constructs
\begin{equation}
    \tilde{x}_t = x_t - s_t.
\end{equation}
Here, $\tilde{x}_t$ denotes the de-seasonalized value provided to the LLM.
This preprocessing step removes dominant recurring patterns before prompting.

However, de-seasonalization does not explicitly represent frequency changes.
It removes a stable seasonal component, but does not describe where local oscillatory behavior changes or how such changes relate to the sequence-level periodic structure.
This limitation motivates adding explicit frequency-domain evidence to the LLM input.

\subsection{Frequency-Domain Representation}
\label{subsec:freq-domain-prelim}

We construct frequency-domain descriptors by applying the Fast Fourier Transform (FFT) to the input time series.
Given $\mathbf{\tilde{x}}=(\tilde{x}_1,\ldots,\tilde{x}_T)$, we compute the frequency-domain descriptors of its mean-centered version.
Let $F_k$ denote the Fourier coefficient at nonzero one-sided frequency bin $k\in\{1,\ldots,K\}$, and define the power spectrum as $P_k=|F_k|^2$.
Excluding the zero-frequency component ensures that the summaries describe oscillatory structure rather than the average signal level.

\noindent\textbf{Dominant frequency and period.}
The dominant frequency is the frequency with the largest spectral power.
Let $k^\star=\arg\max_{1\le k\le K} P_k$.
Then $f_{\mathrm{dom}}=f_{k^\star}$, and the corresponding dominant period is $T_{\mathrm{dom}}=1/f_{\mathrm{dom}}$ when $f_{\mathrm{dom}}>0$.

\noindent\textbf{Spectral entropy.}
Spectral entropy measures whether spectral power is concentrated around a few frequencies or diffusely spread across many frequencies:
\begin{equation}
\begin{gathered}
H
=
-\frac{1}{\log K}
\sum_{k=1}^{K}
p_k \log p_k, \\
p_k
=
\frac{P_k}{\sum_{r=1}^{K} P_r}.
\end{gathered}
\end{equation}
Low entropy indicates concentrated periodic structure, while high entropy indicates a more diffuse spectrum.

\noindent\textbf{Spectral peaks and energy ratio.}
Spectral peaks are the frequency bins with the largest spectral power and summarize the main periodic components beyond a single dominant frequency.
We also compare lower- and higher-frequency energy:
\begin{equation}
    R_{\mathrm{LH}}
    =
    \frac{\sum_{k \in \mathcal{L}} P_k}
         {\sum_{k \in \mathcal{H}} P_k},
\end{equation}
where $\mathcal{L}$ and $\mathcal{H}$ denote the lower and higher halves of the frequency bins excluding the zero-frequency component.
This ratio summarizes whether the spectrum is dominated by slower or faster oscillations.

We call individual FFT-derived quantities, such as dominant period and spectral entropy, frequency-domain descriptors.
A frequency-domain summary consists of descriptors computed at a specific resolution, either global or local.
Once incorporated into the input representation, these summaries serve as frequency-domain evidence rather than anomaly scores or boundary predictions.

\section{Proposed Method}
\begin{figure*}[h]
    \centering
    \includegraphics[width=0.95\textwidth]{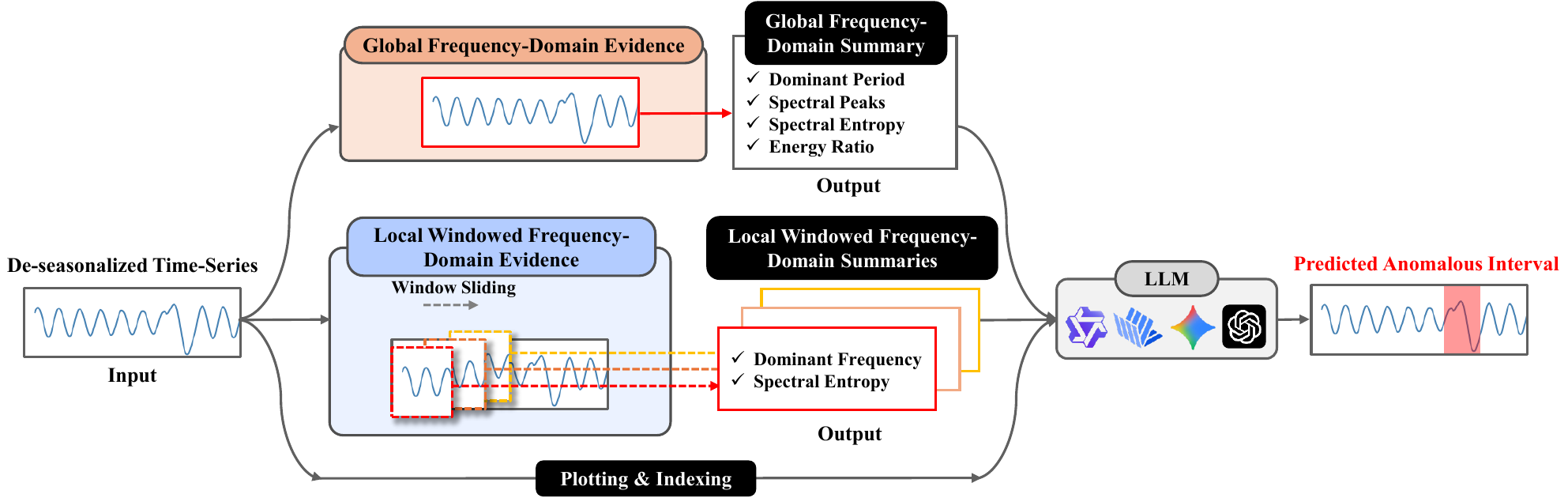}
    \caption{
    \textbf{Overview of the proposed framework.}
    Indexed de-seasonalized values, global frequency-domain evidence, and local windowed frequency-domain evidence are combined in a single prompt for interval prediction.
    }
    \label{fig:main}
    \vspace{-0.3cm}
\end{figure*}

\label{sec:method}

\subsection{Overview}
\label{subsec:method-overview}
We propose an LLM-based time-series anomaly detection framework augmented with explicit frequency-domain evidence.
The framework is based on the observation that indexed time-domain representations, even after de-seasonalization, do not explicitly expose frequency structure to the LLM.
Rather than expecting the LLM to infer periodic structure implicitly from sequence values or plots, we construct explicit frequency-domain evidence from the same de-seasonalized sequence used for interval prediction.

Following LLM-TSAD~\cite{llmtsad}, the time-domain input is represented as an indexed de-seasonalized sequence, denoted by $\tilde{\mathbf{x}}=(\tilde{x}_1,\ldots,\tilde{x}_T)$.
All components in our framework are computed from $\tilde{\mathbf{x}}$.
This design preserves the localization benefit of de-seasonalized indexed representations while adding explicit frequency-domain evidence that is not directly available from the de-seasonalized values alone.

The framework constructs two complementary forms of frequency-domain evidence.
We first compute global frequency-domain evidence to summarize the sequence-level periodic regime.
This global view provides the reference needed to interpret whether a frequency pattern is typical for the full sequence.
We then compute local windowed frequency-domain evidence to expose time-localized spectral departures relative to this sequence-level context.
Thus, global frequency-domain evidence describes the overall periodic reference, while local frequency-domain evidence identifies candidate regions where frequency behavior changes.
Figure~\ref{fig:main} illustrates the overall pipeline.

The final interval prediction is performed from an evidence package containing the indexed de-seasonalized sequence, global frequency-domain evidence, and local frequency-domain evidence.
When visual input is enabled, a time-series plot is additionally included as an auxiliary modality.
These components are not used as standalone anomaly decisions.
Instead, they provide structured support for integrating de-seasonalized time-domain observations, visual cues, and frequency-domain evidence during interval prediction.

\subsection{Global Frequency-Domain Summary}
\label{subsec:global-fft}
Global frequency-domain evidence provides sequence-level periodic context.
It is computed from the indexed de-seasonalized sequence $\tilde{\mathbf{x}}$.
Using the frequency-domain quantities defined in Section~\ref{subsec:freq-domain-prelim}, we compute a global one-sided power spectrum over the full sequence and extract four compact descriptors:
the dominant period, the strongest spectral peaks, global spectral entropy, and the low/high-frequency energy ratio.

The dominant period and strongest peaks summarize the main periodic components of the sequence.
Global spectral entropy measures whether the sequence-level spectrum is concentrated around a few dominant frequencies or distributed across many frequencies.
The low/high-frequency energy ratio summarizes whether the sequence is dominated by slower long-period variation or by stronger high-frequency variation.

The global summary is not associated with temporal location.
Instead, it serves as a sequence-level periodic reference.
This reference is used to interpret whether local spectral departures are unusual relative to the overall periodic regime.

\subsection{Local Windowed Frequency-Domain Summary}
\label{subsec:local-windowed-fft}
Local frequency-domain evidence represents frequency behavior at a time-localized resolution.
A global spectrum describes the overall periodic regime, but it cannot indicate where a local rhythm change occurs.
We therefore divide the indexed de-seasonalized sequence $\tilde{\mathbf{x}}$ into overlapping windows and compute a compact subset of the descriptors for each window.

Given window length $L$ and stride $S$, the $j$-th window is
\begin{equation}
\begin{gathered}
\tilde{\mathbf{w}}^{(j)}
=
(\tilde{x}_{t})_{t=s_j}^{s_j+L-1}, \\
s_j = jS .
\end{gathered}
\end{equation}
for all valid windows satisfying $s_j+L-1<T$.

For each window, we compute the dominant local frequency and local spectral entropy.
Each local window is represented as
\begin{equation}
    \left(s_j,\; s_j+L-1,\; f^{(j)}_{\mathrm{dom}},\; H^{(j)}\right).
\end{equation}
The index range provides coarse temporal support, while the spectral descriptors summarize the local rhythm.
These tuples are used as auxiliary evidence for candidate spectral departures, not as anomaly boundaries.

\subsection{Final Evidence Integration}
\label{subsec:final-evidence-integration}

The final prediction stage integrates the indexed de-seasonalized sequence with the constructed frequency-domain evidence.
For each input sequence, we provide the LLM with an evidence package containing the global and local frequency-domain evidence, and indexed de-seasonalized values.
When visual input is enabled, the corresponding time-series plot is also included.

The global summary provides sequence-level periodic context, while the local summaries provide time-localized spectral cues.
Neither source is treated as a standalone anomaly decision.
Instead, they are used as auxiliary evidence that complements the indexed de-seasonalized sequence, which remains the main basis for interval boundary selection.

The final stage formulates TSAD as interval prediction over the indexed sequence.
The LLM integrates the indexed de-seasonalized values with global and local frequency evidence to produce the final anomalous interval set.
The complete evidence serialization format and output template are provided in Appendix~\ref{app:prompt_templates}.

\section{Experiment}
\label{sec:experiments}
\subsection{Experiment Setup}

\textbf{Datasets.}
We evaluate our method on two benchmarks.
The main evaluation uses AnomLLM~\cite{anomllm}, following the dataset configuration and preprocessing protocol of LLM-TSAD~\cite{llmtsad}.
AnomLLM contains controlled time-series instances with annotated anomaly intervals across representative anomaly types, including point, range, trend, and frequency anomalies.
To further examine whether the proposed evidence representation remains useful beyond this controlled anomaly-type setting, we also evaluate on the same eight-category TSB-AD-U~\cite{tsbadu} evaluation subset used by LLM-TSAD.
This subset-level evaluation is used as an additional practical comparison setting, rather than as exhaustive coverage of the full TSB-AD-U benchmark.

\begin{table*}[t!]
\centering
\setlength{\tabcolsep}{10pt}
\renewcommand{\arraystretch}{1}

\resizebox{0.9\textwidth}{!}{%
\begin{tabular}{ll ccc ccc}
\toprule
\multirow{2}{*}{\textbf{LLMs}} & \multirow{2}{*}{\textbf{Methods}} & \multicolumn{3}{c}{\textbf{Standard Metrics}} & \multicolumn{3}{c}{\textbf{Affiliation Metrics}} \\
\cmidrule(lr){3-5} \cmidrule(lr){6-8}
& & \textbf{Prec.} & \textbf{Recall} & \textbf{F1} & \textbf{Prec.} & \textbf{Recall} & \textbf{F1} \\
\midrule

\multicolumn{2}{l}{\textit{Na\"ive baseline (Always Normal Predictor)}} 
& 31.75 & 31.75 & 31.75 
& 31.75 & 31.75 & 31.75 \\
\midrule

\multirow{8}{*}{\shortstack[l]{\textbf{InternVL2}\\[-1pt]\scriptsize (LLaMA3-76B)}} 
& \multicolumn{7}{l}{\textit{Existing Prompts (AnomLLM)}} \\
& \hspace{3mm} 0shot-Text 
& 12.94 & 21.15 & 13.38 
& 22.12 & 28.38 & 24.10 \\
& \hspace{3mm} 0shot-Vision 
& 22.33 & 46.19 & 23.92 
& 50.93 & 60.94 & 55.53 \\
& \hspace{3mm} 1shot-Vision-CoT 
& \underline{33.72} & 36.28 & 33.67 
& 53.62 & 55.55 & 53.93 \\
\cmidrule(lr){2-8}
& LLM-TSAD (0shot-Text) 
& \underline{27.67} & \underline{52.29} & \underline{29.21} 
& \underline{54.31} & \underline{60.71} & \underline{55.04} \\
& LLM-TSAD (0shot-Text+Vision) 
& 30.00 & \underline{59.35} & \underline{34.66} 
& \underline{58.14} & \underline{71.79} & \underline{62.90} \\
\cmidrule(lr){2-8}
\rowcolor{gray!10} & \textbf{Ours (0shot-Text)} 
& \textbf{44.66} & \textbf{66.83} & \textbf{46.64} 
& \textbf{73.14} & \textbf{74.01} & \textbf{71.93} \\
\rowcolor{gray!10} & \textbf{Ours (0shot-Text+Vision)} 
& \textbf{36.99} & \textbf{72.28} & \textbf{42.15} 
& \textbf{69.54} & \textbf{78.45} & \textbf{72.41} \\

\midrule

\multirow{8}{*}{\shortstack[l]{\textbf{Qwen-2.5}\\[-1pt]\scriptsize (VL-72B-Instruct)}} 
& \multicolumn{7}{l}{\textit{Existing Prompts (AnomLLM)}} \\
& \hspace{3mm} 0shot-Text 
& 25.99 & 25.66 & 25.49 
& 45.69 & 42.44 & 43.11 \\
& \hspace{3mm} 0shot-Vision 
& 45.78 & 51.00 & 46.65 
& 69.50 & 68.99 & 68.78 \\
& \hspace{3mm} 1shot-Vision-CoT 
& 27.09 & 29.08 & 27.23 
& 39.52 & 40.07 & 39.38 \\
\cmidrule(lr){2-8}
& LLM-TSAD (0shot-Text) 
& \textbf{72.47} & \underline{54.09} & \underline{56.13} 
& \underline{82.91} & \underline{78.40} & \underline{78.96} \\
& LLM-TSAD (0shot-Text+Vision) 
& \underline{72.95} & \underline{69.78} & \underline{67.56} 
& \underline{84.51} & \underline{81.52} & \underline{81.87} \\
\cmidrule(lr){2-8}
\rowcolor{gray!10} & \textbf{Ours (0shot-Text)} 
& \underline{68.80} & \textbf{68.54} & \textbf{64.38} 
& \textbf{87.22} & \textbf{85.44} & \textbf{85.39} \\
\rowcolor{gray!10} & \textbf{Ours (0shot-Text+Vision)} 
& \textbf{77.93} & \textbf{77.20} & \textbf{73.70} 
& \textbf{93.88} & \textbf{90.64} & \textbf{91.39} \\

\midrule

\multirow{8}{*}{\shortstack[l]{\textbf{Gemini-2.5}\\[-1pt]\scriptsize (Flash)}} 
& \multicolumn{7}{l}{\textit{Existing Prompts (AnomLLM)}} \\
& \hspace{3mm} 0shot-Text 
& 15.69 & 12.83 & 13.37 
& 42.64 & 38.31 & 38.85 \\
& \hspace{3mm} 0shot-Vision 
& 52.07 & 70.67 & 57.28 
& 81.39 & 79.30 & 79.55 \\
& \hspace{3mm} 1shot-Vision-CoT 
& 46.64 & 61.54 & 50.96 
& 75.51 & 74.80 & 74.55 \\
\cmidrule(lr){2-8}
& LLM-TSAD (0shot-Text) 
& \textbf{69.54} & \underline{61.95} & \underline{62.87} 
& \underline{77.21} & \underline{75.43} & \underline{75.58} \\
& LLM-TSAD (0shot-Text+Vision) 
& \textbf{80.88} & \underline{76.44} & \underline{76.62} 
& \underline{88.55} & \underline{86.90} & \underline{87.12} \\
\cmidrule(lr){2-8}
\rowcolor{gray!10} & \textbf{Ours (0shot-Text)} 
& \underline{65.91} & \textbf{72.22} & \textbf{64.38} 
& \textbf{82.20} & \textbf{82.58} & \textbf{82.02} \\
\rowcolor{gray!10} & \textbf{Ours (0shot-Text+Vision)} 
& \underline{78.48} & \textbf{84.72} & \textbf{77.97} 
& \textbf{92.66} & \textbf{92.80} & \textbf{92.48} \\

\midrule

\multirow{8}{*}{\textbf{GPT-4o}} 
& \multicolumn{7}{l}{\textit{Existing Prompts (AnomLLM)}} \\
& \hspace{3mm} 0shot-Text 
& 19.69 & 17.62 & 17.76 
& 46.35 & 45.51 & 44.73 \\
& \hspace{3mm} 0shot-Vision 
& 39.54 & 50.60 & 42.03 
& 62.19 & 61.99 & 61.68 \\
& \hspace{3mm} 1shot-Vision-CoT 
& 31.48 & 38.50 & 32.93 
& 56.43 & 55.84 & 55.39 \\
\cmidrule(lr){2-8}
& LLM-TSAD (0shot-Text) 
& \textbf{65.52} & \underline{53.45} & \underline{54.40} 
& \underline{75.65} & \underline{73.88} & \underline{73.47} \\
& LLM-TSAD (0shot-Text+Vision) 
& \textbf{76.94} & \underline{68.42} & \underline{70.04} 
& \underline{83.10} & \underline{78.85} & \underline{80.11} \\
\cmidrule(lr){2-8}
\rowcolor{gray!10} & \textbf{Ours (0shot-Text)} 
& \underline{58.65} & \textbf{63.79} & \textbf{55.51} 
& \textbf{79.76} & \textbf{78.73} & \textbf{78.56} \\
\rowcolor{gray!10} & \textbf{Ours (0shot-Text+Vision)} 
& \underline{72.30} & \textbf{75.75} & \textbf{70.20} 
& \textbf{89.51} & \textbf{86.71} & \textbf{87.35} \\

\bottomrule
\end{tabular}%
}
\vspace{-0.1cm}
\caption{
\textbf{Main anomaly detection performance on AnomLLM across four multimodal LLMs.}
We report standard and affiliation-based precision, recall, and F1.
For each model block and input modality group, the best result is shown in \textbf{bold}, and the second-best result is \underline{underlined}.
}
\label{tab:main}
\vspace{-0.3cm}
\end{table*}

\noindent\textbf{Evaluation models.}
We evaluate the proposed method with four vision-language models: InternVL2-LLaMA3-76B~\cite{intern}, Qwen2.5-VL-72B-Instruct~\cite{qwen}, Gemini-2.5-Flash~\cite{gemini}, and GPT-4o~\cite{gpt4o}.
These models include both open-source and proprietary multimodal LLMs, allowing us to test whether the proposed evidence representation is useful across different model families.
For the TSB-AD-U evaluation subset, we use Gemini-2.5-Flash to compare our method with Gemini-based LLM baselines under the same backbone.

\noindent\textbf{Baselines.}
On AnomLLM, we compare with representative AnomLLM prompting variants~\cite{anomllm} and LLM-TSAD~\cite{llmtsad}, our main baseline.
LLM-TSAD is most relevant because our method preserves its indexed de-seasonalized formulation while adding FFT-based global and local frequency-domain evidence.
We also report the na\"ive always-normal predictor as a dataset-level prior.

For the TSB-AD-U subset, we follow the compact LLM-TSAD protocol and report the strongest representative non-LLM baseline from each family, together with Gemini-2.5-Flash-based LLM baselines.

\noindent\textbf{Metrics.}
Following LLM-TSAD~\cite{llmtsad}, we report both standard metrics and affiliation metrics.
For each metric family, we report precision, recall, and F1.
Standard metrics evaluate label-level overlap between predicted and ground-truth anomaly intervals, providing a strict measure of interval prediction accuracy.
Affiliation metrics provide a complementary distance-aware assessment of temporal agreement between predicted and ground-truth intervals.
All scores are aggregated over evaluation instances following the same averaging protocol as LLM-TSAD~\cite{llmtsad}.
\begin{figure*}[h!]
    \centering
    \includegraphics[width=0.9\textwidth]{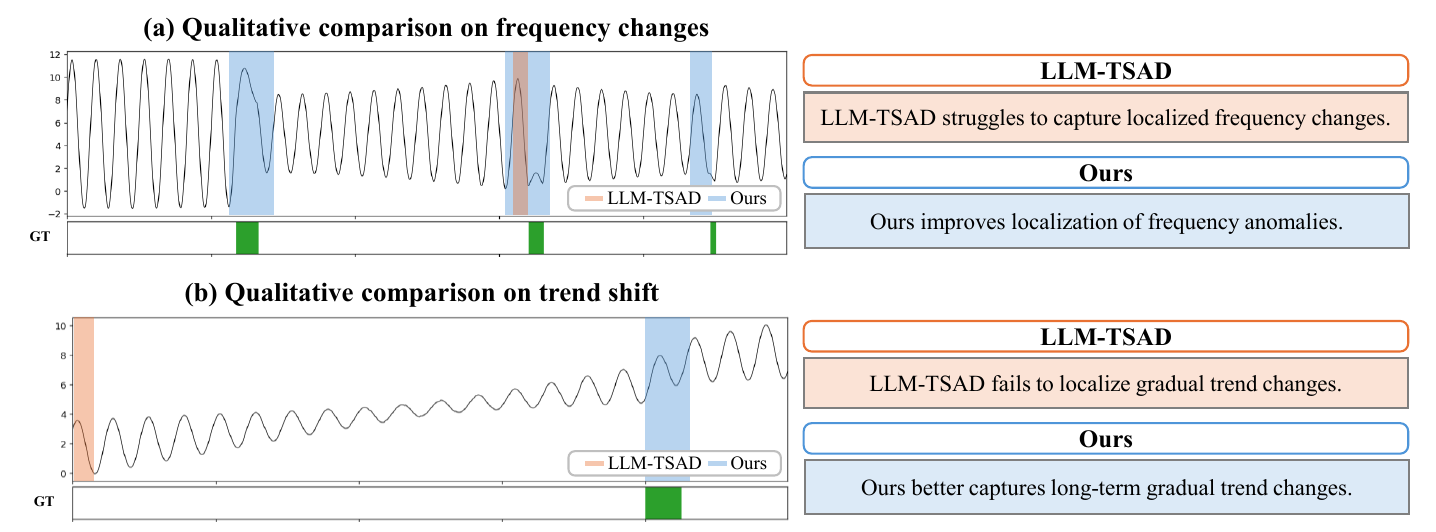}
    \caption{
    \textbf{Qualitative comparison between LLM-TSAD and our method using Qwen2.5-VL-72B-Instruct.}
    The examples show frequency-change and trend-shift anomalies from AnomLLM.
    }
    \label{fig:qual_result}
    \vspace{-0.2cm}
\end{figure*}

\subsection{Main Results}
\label{subsec:main-results}

\textbf{Quantitative results.}
Table~\ref{tab:main} reports the main results on AnomLLM across four multimodal LLMs.
Compared with LLM-TSAD, our method consistently improves both standard F1 and affiliation F1 across all four evaluated models and both input settings.
These results show that the proposed frequency-domain evidence improves both strict interval overlap and distance-aware temporal agreement.

The magnitude of improvement varies across backbone and modality configurations, reflecting differences in how multimodal LLMs integrate indexed numerical values, visual plots, and structured frequency-domain evidence.
For InternVL2, the text-only setting achieves the strongest standard F1, while the text+vision setting yields the highest affiliation F1 within the same backbone.
This pattern suggests that visual context does not uniformly affect strict boundary overlap, but can still support distance-aware temporal agreement under affiliation-based evaluation.
Overall, frequency-domain summaries act as complementary evidence for LLM-based TSAD, with backbone-specific modality effects.

\begin{table}[t]
\centering
\footnotesize
\setlength{\tabcolsep}{7pt}
\renewcommand{\arraystretch}{1}

\begin{tabular}{llcc}
\toprule
\textbf{Model}
& \textbf{Setting}
& \textbf{F1}
& \textbf{Affi. F1} \\
\midrule

\multirow{3}{*}{InternVL2}
& LLM-TSAD
& 34.66 & 62.90 \\
\cmidrule(lr){2-4}
& + LF
& \underline{40.31} & \underline{72.27} \\
& + LF + GF
& \textbf{42.15} & \textbf{72.41} \\

\midrule

\multirow{3}{*}{Qwen-2.5}
& LLM-TSAD
& 67.56 & 81.87 \\
\cmidrule(lr){2-4}
& + LF
& \underline{71.62} & \textbf{92.57} \\
& + LF + GF
& \textbf{73.70} & \underline{91.39} \\

\midrule

\multirow{3}{*}{Gemini-2.5}
& LLM-TSAD
& \underline{76.62} & \underline{87.12} \\
\cmidrule(lr){2-4}
& + LF
& 62.86 & 80.44 \\
& + LF + GF
& \textbf{77.97} & \textbf{92.48} \\

\midrule

\multirow{3}{*}{GPT-4o}
& LLM-TSAD
& \underline{70.04} & 80.11 \\
\cmidrule(lr){2-4}
& + LF
& 67.40 & \textbf{89.02} \\
& + LF + GF
& \textbf{70.20} & \underline{87.35} \\

\bottomrule
\end{tabular}
\caption{
\textbf{Component-to-final ablation.}
LF and GF denote local and global frequency-domain evidence, respectively, with LF+GF representing the final frequency-augmented configuration.
The best and second-best results within each model are shown in \textbf{bold} and \underline{underlined}.
}
\vspace{-0.5cm}
\label{tab:component-to-final}
\end{table}

\noindent\textbf{Qualitative results.}
Figure~\ref{fig:qual_result} shows representative cases where anomalies involve localized rhythm changes or gradual temporal structure shifts rather than isolated pointwise deviations.
Compared with LLM-TSAD, our method produces intervals that better align with the ground-truth regions, illustrating how explicit frequency-domain evidence can complement indexed and de-seasonalized time-domain inputs.
These examples support the interpretation of the quantitative results, but are used only as qualitative evidence rather than standalone proof of overall superiority.

\subsection{Ablation Studies}
\begin{table*}[t]
\centering
\setlength{\tabcolsep}{12pt}
\renewcommand{\arraystretch}{1.1}

\resizebox{1\textwidth}{!}{%

\begin{tabular}{l|l|ccc|ccc}
\toprule
\multirow{2}{*}{\textbf{Group}}
& \multirow{2}{*}{\textbf{Method}}
& \multicolumn{3}{c|}{\textbf{Standard Metrics}}
& \multicolumn{3}{c}{\textbf{Affiliation Metrics}} \\
\cmidrule(lr){3-5} \cmidrule(lr){6-8}
&
& \textbf{Prec.}
& \textbf{Recall}
& \textbf{F1}
& \textbf{Prec.}
& \textbf{Recall}
& \textbf{F1} \\
\midrule

\multirow{3}{*}{Non-LLM baseline}
& ML: SR~\cite{sr}
& 32.61 & 40.77 & 30.57
& 66.80 & \textbf{95.11} & \underline{75.94} \\

& DL: USAD~\cite{usad}
& 23.92 & 34.54 & 24.46
& 59.84 & 61.60 & 55.24 \\

& FM: Chronos~\cite{ansari2024chronos}
& 23.58 & \textbf{55.65} & 25.08
& 64.87 & \underline{96.33} & 75.90 \\

\midrule

\multirow{2}{*}{LLM baseline}
& AnomLLM (Gemini-2.5-Flash)
& 12.28 & 16.93 & 6.82
& 40.37 & 33.75 & 30.97 \\

& LLM-TSAD (Gemini-2.5-Flash)
& \underline{43.05} & 43.49 & \underline{36.85}
& \underline{73.08} & 93.88 & 80.15 \\

\midrule

\rowcolor{gray!10}
\textbf{Ours}
& \textbf{Ours (Gemini-2.5-Flash)}
& \textbf{46.51} & \underline{46.19} & \textbf{40.26}
& \textbf{74.62} & 92.71 & \textbf{80.35} \\

\bottomrule
\end{tabular}
}
\caption{
\textbf{Overall performance on the eight-category TSB-AD-U subset used by LLM-TSAD.}
We compare our method with representative non-LLM and Gemini-based LLM baselines.
}
\label{tab:tsb_overall}
\end{table*}

\begin{table}[t]
\centering
\footnotesize
\setlength{\tabcolsep}{4pt}
\renewcommand{\arraystretch}{1.50}

\begin{tabular}{l|cccc|c}
\toprule
\textbf{Config.}
& Freq.
& Point
& Range
& Trend
& \textbf{Overall} \\
\midrule

Indexed
& 23.99
& \textbf{88.13}
& \textbf{86.39}
& 60.28
& 64.70 \\

+D.S.$^\dagger$
& 33.98
& 77.51
& \underline{85.58}
& 71.51
& 67.15 \\

\midrule

+D.S.+LF
& \underline{39.24}
& 81.55
& 82.85
& \underline{82.83}
& \underline{71.62} \\

+D.S.+LF+GF
& \textbf{40.89}
& \underline{83.30}
& 83.94
& \textbf{86.67}
& \textbf{73.70} \\

\bottomrule
\end{tabular}
\caption{
\textbf{Type-wise standard F1 across evidence configurations.}
We report standard F1 by anomaly type using Qwen2.5-VL-72B-Instruct.
D.S., LF, and GF denote de-seasonalization, local frequency-domain evidence, and global frequency-domain evidence, respectively.
The best and second-best results for each anomaly type are shown in \textbf{bold} and \underline{underlined}, respectively.
$\dagger$ denotes our reproduced result.
}
\label{tab:type-wise-analysis}
\end{table}

\textbf{Component-to-final ablation.}
Table~\ref{tab:component-to-final} evaluates how local frequency-domain evidence (LF) and global frequency-domain evidence (GF) contribute to the final prediction.
LF-only is an intermediate configuration that exposes window-level spectral departures without sequence-level periodic context.
While LF improves affiliation F1 for most models, the Gemini-2.5 result indicates that local evidence alone can be insufficient or less reliable for stable interval localization.
Adding GF on top of LF improves standard F1 for all models and restores or further improves affiliation F1 in several cases, suggesting that global periodic context helps calibrate local spectral departures.
Overall, the ablation supports the use of LF and GF as complementary evidence in the final configuration.

\noindent\textbf{Type-wise analysis.}
Table~\ref{tab:type-wise-analysis} breaks down standard F1 by anomaly type.
The largest gains appear for frequency changes and trend shifts, where anomalies reflect changes in rhythm, periodicity, or long-term temporal structure rather than isolated value deviations.
Although trend shifts are not frequency anomalies in the strict sense, they can induce low-frequency energy changes or alter local spectral distributions, making them partially observable through FFT-based local and global summaries.
This aligns with the roles of LF and GF: local evidence captures window-level spectral changes, while global evidence provides sequence-level periodic context.

For point and range anomalies, the indexed de-seasonalized sequence already exposes strong amplitude- and boundary-level cues, so frequency-domain summaries provide less additional information.
Overall, the type-wise results support a targeted interpretation of our method: frequency-domain evidence is most useful when anomaly localization requires spectral or long-range temporal context beyond local time-domain values.

\noindent\textbf{Evaluation on TSB-AD-U.}
Table~\ref{tab:tsb_overall} reports results on the same eight-category TSB-AD-U subset used by LLM-TSAD: NEK, TAO, MSL, Power, Daphnet, YAHOO, SED, and TODS.
This setting follows the prior subset-level protocol to provide a cross-benchmark robustness check against the strongest available LLM-based TSAD baseline, rather than an exhaustive evaluation of the full benchmark.

With the same Gemini-2.5-Flash backbone, our method improves standard F1 over Gemini-based LLM-TSAD while maintaining comparable affiliation-based performance.
This supports the usefulness of the proposed frequency-domain summaries beyond AnomLLM, within the scope of the subset-level evaluation.

\section{Conclusion}
We proposed an evidence-augmented framework for zero-shot LLM-based time-series anomaly detection that complements indexed and de-seasonalized time-domain inputs with compact global and local frequency-domain evidence.
Global evidence provides sequence-level periodic context, while local evidence exposes time-localized spectral departures; both are used as auxiliary evidence rather than standalone anomaly decisions.
Our probing experiments suggest that such frequency-relevant structure is not always reliably recovered from indexed values or plots alone.
Experiments on AnomLLM with multiple multimodal LLMs and evaluation on the TSB-AD-U evaluation subset show that explicit frequency-domain evidence can improve strong LLM-based baselines, especially for anomalies involving rhythm, periodicity, or gradual temporal changes.
These findings motivate adaptive frequency-evidence construction for longer or multivariate time series.
\clearpage

\section{Limitations}
\label{app:limitations}
While frequency-domain evidence improves LLM-based TSAD in our experiments, zero-shot anomaly localization remains challenging.
We do not directly analyze which parts of the provided evidence the LLM relies on when producing its interval predictions.
Therefore, our results support the usefulness of structured frequency-domain evidence at the input level, but do not fully explain the internal reasoning process of the model.
Future work may investigate attribution or intervention-based analyses to better understand how LLMs use time-domain, visual, and frequency-domain evidence during anomaly localization.

\bibliography{custom}

\clearpage
\appendix

\label{sec:appendix}
\section*{Table of Contents}
\begin{description}
    \item[\textbf{A.}] Related Work \dotfill \pageref{app:related}
    \item[\textbf{B.}] Computational Resources and Experimental Setup \dotfill \pageref{app:resources}
    \item[\textbf{C.}] Prompt Templates \dotfill \pageref{app:prompt_templates}
    \item[\textbf{D.}] Frequency-Reasoning Probe Setup \dotfill \pageref{app:probe-setup}
    \item[\textbf{E.}] Implementation Details \dotfill \pageref{app:i.d.}
    \item[\textbf{F.}] Additional Experimental Analyses \dotfill \pageref{app:additional-analyses}
    \item[\textbf{G.}] Prompt Overhead and Evaluation Details \dotfill \pageref{app:eval-details}
\end{description}

\section{Related Work}
\label{app:related}

Time-series anomaly detection (TSAD) commonly uses reconstruction, forecasting, or representation-learning objectives. Classic paradigms include statistical decomposition like STL~\cite{cleveland1990stl}, decomposable forecasting models such as Prophet~\cite{taylor2018prophet}, and bio-inspired online learning like HTM~\cite{hawkins2010htm}. Additionally, algorithmic approaches such as Matrix Profiles~\cite{yeh2018matrix} and ensemble-based HS-Trees~\cite{tan2011fast} capture complex sequence representations. Deep learning models further extend this paradigm by modeling probabilistic latent structures, adversarial reconstruction errors, temporal dependencies, and frequency-aware representations~\cite{su2019robust, usad, xu2021anomaly, tfad, wu2025temporal, Yi2023Frequency}. Related efforts to characterize normality also appear in visual anomaly detection, including density-based and non-conformity-based formulations~\cite{seo2026anomaly, kim2023sanflow}. Within TSAD, however, most conventional approaches remain dataset-specific and typically require sufficient anomaly-free training data, which limits their applicability in dynamic or previously unseen environments.

Recent progress in foundation models motivates a shift from task-specific TSAD models toward more generalizable approaches. Large-scale pre-trained time-series models show that generic sequence modeling can support numerical forecasting across diverse domains~\cite{ansari2024chronos, das2024decoder}. In parallel, LLM-based approaches examine whether time-series signals can be transformed into textual representations and analyzed through in-context reasoning. By using textualization, prompting, and retrieval-based demonstrations, these methods aim to support zero-shot anomaly detection while also producing interpretable explanations~\cite{alnegheimish2024can, llmad}.

Although LLM-based methods improve flexibility and interpretability, applying language models directly to continuous numerical signals remains challenging~\cite{jin2024time}. Since LLMs are primarily trained on discrete semantic tokens, naively serialized time-series inputs may not preserve fine-grained local variations, periodic structures, or subtle frequency changes in a form that the model can reliably interpret~\cite{anomllm, zheng2025understanding}. As a result, existing prompting-based detectors may become sensitive to visually salient amplitude changes while being less reliable for contextual anomalies, phase shifts, or frequency-domain irregularities~\cite{dualtf}. Moreover, when anomaly localization relies only on textual indices, the model must implicitly track positions and compare numerical values, which can impose an additional reasoning burden on the LLM.

Recent work mitigates these issues by combining statistical preprocessing with structured prompting. For example, de-seasonalization can simplify the input signal and expose residual deviations\cite{xue2022PromptCast, liu2023unitime}, while index-aware prompting provides explicit positional information~\cite{llmtsad}. However, suppressing seasonal components is not the same as explicitly representing frequency structure. Although de-seasonalization can help localization, it does not directly describe how local oscillatory behavior changes or how such changes relate to the sequence-level periodic regime. This makes preprocessing-based simplification incomplete for anomalies expressed through rhythm, periodicity, or transient frequency changes.

Our evidence-augmented framework addresses this limitation by preserving the indexed time-domain signal while adding compact frequency-domain evidence.
Local frequency-domain summaries expose time-localized spectral departures, while global frequency-domain summaries provide sequence-level periodic context.
Together, these components form a structured interface that allows zero-shot LLM-based TSAD to use both time-domain and frequency-domain evidence.

\section{Computational Resources and Experimental Setup}
\label{app:resources}

In this section, we provide details regarding the hardware and software environments for preprocessing, model inference, and server execution.

\subsection{Computing Infrastructure}
Most preprocessing steps, including data cleaning, de-seasonalization, and frequency-domain evidence construction, run on a local workstation. The hardware specifications are as follows:
\begin{itemize}
    \item \textbf{CPU:} Intel(R) Core(TM) i9-10980XE CPU @ 3.00GHz (18 cores, 36 threads).
    \item \textbf{GPU:} NVIDIA GeForce RTX 4090 (24GB VRAM).
    \item \textbf{Usage:} This infrastructure supports dataset preprocessing, FFT-based evidence construction, and the multimodal inference pipeline.
\end{itemize}

For the InternVL2~\cite{intern} evaluation, we use one NVIDIA B200 GPU through the Runpod cloud service.
In this setting, the corresponding FFT-based evidence construction and model inference run on the same cloud server.

\subsection{API-based Model Specifications}

For API-based LLM inference, we use the model endpoints listed in Table~\ref{tab:api_models}.
The inference pipeline accesses these endpoints through OpenRouter.

\begin{table}[t]
\centering
\scriptsize
\begin{tabular}{lll}
\toprule
Provider & Model & Endpoint identifier \\
\midrule
OpenAI & GPT-4o & \texttt{openai/gpt-4o} \\
Google & Gemini 2.5 Flash & \texttt{google/gemini-2.5-flash} \\
Qwen & Qwen 2.5 VL & \texttt{qwen/qwen2.5-vl-72b-instruct} \\
\bottomrule
\end{tabular}
\caption{
API-based model endpoints for our experiments.
}
\label{tab:api_models}
\end{table}

\subsection{Artifact Use}
We use AnomLLM~\cite{anomllm} and the TSB-AD-U~\cite{tsbadu} subset only for research evaluation, consistent with their intended use as anomaly-detection benchmarks. 
We access API-based and open-source LLMs only through their documented inference interfaces.
Derived preprocessing outputs, including de-seasonalized sequences and FFT-based evidence summaries, support only anomaly-detection evaluation and are not redistributed.

\subsection{Data Privacy and Content}
We use publicly available time-series anomaly detection benchmarks, AnomLLM and the TSB-AD-U subset, only for research evaluation. The data used in our experiments consist of numerical time-series values and anomaly interval annotations rather than natural-language user content. We do not identify personally identifying information or offensive textual content in the benchmark inputs for evaluation. We collect no additional personal data and make no attempt to re-identify individuals or link the data to external sources.

\subsection{Data Statistics}
We use the AnomLLM benchmark following the dataset configuration and preprocessing protocol of LLM-TSAD. We also evaluate on the eight-category TSB-AD-U subset used by LLM-TSAD, consisting of NEK, TAO, MSL, Power, Daphnet, YAHOO, SED, and TODS. Since our method is evaluated in a zero-shot setting, no train/dev split is used for model training or prompt tuning.

\section{Prompt Templates}
\label{app:prompt_templates}

We report the prompt template for our frequency-augmented LLM-based TSAD framework.
The full template is shown in Figure~\ref{fig:prompt_template}.
Ellipses (\texttt{...}) indicate omitted consecutive numeric entries for readability; in the actual prompt, these entries are provided explicitly.
For prompt serialization, we use zero-based indices to match the benchmark input format, although the mathematical notation in the main text follows the conventional one-based form.
Following the LLM-TSAD~\cite{llmtsad} evaluation protocol, the prompt specifies that each sequence contains up to five anomalous intervals.
This constraint is used only to match the prior benchmark protocol and is applied consistently across the LLM-based methods in the comparison; no anomaly-type labels or instance-specific demonstrations are provided during inference.
For text-only settings, the image-related instruction is omitted from the prompt, while the remaining evidence serialization and output format are kept unchanged.

\begin{figure*}[t]
\centering
\begin{lstlisting}[style=promptstyle]
I will provide you with time-series value data recorded at hourly intervals, along with a plotted time-series image.

<global_frequency_analysis>
The following global frequency-domain features summarize the overall frequency structure of the entire time series.

Use global frequency-domain evidence to understand the baseline periodic structure of the sequence.
Do not use global frequency-domain evidence alone to determine the anomaly interval.

Dominant frequency: 0.0310 (period ~= 32.3 steps)
Top spectral peaks: 0.0310, 0.0300, 0.0290
Spectral entropy (normalized): 0.4374
Low/High frequency energy ratio: 1149.75
</global_frequency_analysis>

<local_frequency_analysis>
The following local/windowed frequency-domain features summarize frequency behavior across time windows.

Use local frequency-domain evidence to identify where frequency behavior changes within the sequence.
Localized changes in dominant frequency or spectral entropy may support point, burst, oscillatory, or frequency anomalies.

Window [0-63]: dominant_freq=0.0312, spectral_entropy=0.1295
Window [32-95]: dominant_freq=0.0312, spectral_entropy=0.1400
...
Window [864-927]: dominant_freq=0.0312, spectral_entropy=0.0958
Window [896-959]: dominant_freq=0.0312, spectral_entropy=0.0913
</local_frequency_analysis>

Here is time-series data in (index, value) format:
<history>
(0, 7.13)
(1, 7.52)
...
(998, -0.77)
(999, -0.14)
</history>

Assume there are up to 5 anomalies.

Detect ranges of anomalies in this time series, considering the plotted image if it is useful.
The index of the time series starts from 0 to 999.
List one by one, in JSON format.
If there are no anomalies, answer with an empty list [].
Do not say anything other than the answer.

Output template:
[{"start": ..., "end": ...}, {"start": ..., "end": ...}, ...]
\end{lstlisting}
\caption{
Prompt template used in the proposed frequency-augmented LLM-based TSAD framework.
}
\label{fig:prompt_template}
\end{figure*}

\section{Frequency-Reasoning Probe Setup}
\label{app:probe-setup}

We use three synthetic probing experiments to examine whether different input representations expose frequency-relevant information to an LLM.
These probes serve as diagnostic representation tests that motivate explicit frequency-domain evidence rather than full TSAD benchmarks.

\subsection{Common Setup}
Each probe uses 200 synthetic samples.
All evaluations use Qwen2.5-VL-72B-Instruct~\cite{qwen}.
For each sample, we compare five input representations: indexed text, indexed text with a plot, de-seasonalized text, indexed text augmented with global frequency-domain evidence, and indexed text augmented with local frequency-domain evidence.
Global frequency-domain evidence comes from a frequency-domain summary over the full sequence.
Local frequency-domain evidence comes from window-level frequency-domain summaries with sliding windows of length 64 and stride 32.
When constructing frequency-domain summaries, we exclude periods shorter than 4 samples or longer than 0.8 times the analyzed sequence or window length.

\subsection{Single-Sinusoid Probe}
Each sequence follows
\[
x_t = A \sin(2\pi t / P + \phi) + \epsilon_t,
\]
where we sample the sequence length from \(\{128,256,512\}\), the period \(P\) from \(\{8,12,16,24,32,48,64\}\), the amplitude \(A\) from \([0.5,2.0]\), and the Gaussian noise standard deviation from \([0.02,0.15]\).
The model outputs the dominant period.
We report Top-1 Recovery with a 10\% relative-error tolerance, where a prediction is counted as correct if
\[
|\hat{P}-P|/P \leq 0.10.
\]

\subsection{Multi-Sinusoid Probe}
Each sequence follows a sum of 2 to 4 sinusoidal components:
\[
x_t = \sum_i A_i \sin(2\pi t / P_i + \phi_i) + \epsilon_t.
\]
We sample the sequence length from \(\{256,512\}\) and component periods without replacement from \(\{8,12,16,24,32,48,64\}\).
Amplitude and noise ranges follow the single-sinusoid probe.
The model outputs the dominant period and the top-\(k\) periods.
We report Top-\(k\) Recovery with a 10\% relative-error tolerance, computed as the fraction of true periods matched by any predicted period and then macro-averaged over samples.

\subsection{Local Frequency-Change Probe}
Each sequence contains one contiguous segment whose frequency differs from the background while the amplitude range remains comparable:
\[
x_t =
A \sin(2\pi t / P_s + \phi_s) + \epsilon_t,
\]
where \((P_s,\phi_s)=(P_{\mathrm{bg}},\phi_{\mathrm{bg}})\) outside the changed-frequency segment and \((P_s,\phi_s)=(P_{\mathrm{anom}},\phi_{\mathrm{anom}})\) inside the segment.
We sample the sequence length from \(\{256,512\}\) and the background period from \(\{16,24,32\}\). We set the anomalous period to \(P_{\mathrm{bg}}r\), where \(r \in \{0.4,0.5,2.0,3.0\}\), with clipping to a valid range.
We sample the changed-frequency segment length from 20 to 80 time steps and place it away from the first and last 10\% of the sequence.
We sample amplitude from \([0.8,1.2]\) and the Gaussian noise standard deviation from \([0.02,0.10]\).
The model determines whether the sequence contains a local frequency-change segment.
The reported metric is binary detection accuracy, computed from the predicted anomaly-presence flag.

\subsection{De-seasonalization}
For the de-seasonalized representation, we construct a seasonal template from an oracle period and subtract the tiled template from the original sequence.
The oracle period serves only to define a strong diagnostic de-seasonalized representation.
For the local frequency-change probe, this period is the background period.
For the single- and multi-sinusoid probes, it is the ground-truth dominant period.
This oracle-period construction is used only in the synthetic probing experiments and is not used in the AnomLLM~\cite{anomllm}or TSB-AD-U~\cite{tsbadu} benchmark evaluations.

\section{Implementation Details}
\label{app:i.d.}
For local windowed frequency-domain evidence, we use a window length of 64 for the AnomLLM benchmark~\cite{anomllm} and 128 for the TSB-AD-U subset~\cite{tsbadu}.
Since the input sequences in the TSB-AD-U subset are approximately twice as long as those in AnomLLM under our evaluation setting, we double the window length to keep the relative temporal scale of local windowed frequency-domain summaries comparable across benchmarks. 
In both benchmarks, the window stride is set to half of the corresponding window length. 
For the global frequency-domain summary, we retain the top three nonzero-frequency spectral peaks with the largest spectral power across all experiments.

\section{Additional Experimental Analyses}
\label{app:additional-analyses}

This section consolidates the controlled analyses that complement the main evaluation. The analyses target three questions that follow directly from the motivation of the paper: whether raw FFT descriptors alone explain the gains, whether global and local descriptors play distinct roles, and whether the observed behavior remains stable under evidence perturbations, temporal-resolution changes, and backbone changes.

\subsection{Controlled Diagnostic Subset}
\label{app:controlled-subset}
Several analyses below use a balanced 1/4-scale subset of AnomLLM. This subset serves as a controlled diagnostic setting for comparisons that require many matched LLM calls. Every condition uses the same instances and preserves anomaly-type balance, which isolates the changed evidence representation while keeping repeated diagnostic inference tractable. The subset-scale analyses do not replace the full-scale results in the main paper; they test narrower properties of the representation under matched conditions.

\subsection{Structured Evidence versus Naive Descriptor Serialization}
\label{app:prompt-schema-ablation}
\begin{table}[t]
\centering
\scriptsize
\setlength{\tabcolsep}{3pt}
\begin{tabularx}{\columnwidth}{@{}lXrr@{}}
\toprule
Model & Configuration & F1 & Affi. F1 \\
\midrule
Qwen2.5-VL-72B & Baseline, no frequency descriptors & 68.33 & 82.78 \\
Qwen2.5-VL-72B & Naive descriptor serialization & 67.24 & 85.62 \\
Qwen2.5-VL-72B & Structured GF/LF evidence & \textbf{73.97} & \textbf{92.53} \\
\midrule
Gemini-2.5-Flash & Baseline, no frequency descriptors & 75.84 & 86.76 \\
Gemini-2.5-Flash & Naive descriptor serialization & 66.24 & 83.99 \\
Gemini-2.5-Flash & Structured GF/LF evidence & \textbf{77.27} & \textbf{92.09} \\
\bottomrule
\end{tabularx}
\caption{Same-LLM prompt-schema ablation on the balanced 1/4-scale diagnostic subset. All conditions keep the LLM, instances, image, indexed history, output format, descriptor values and order, and block position fixed; only the representation schema changes.}
\label{tab:appendix-prompt-schema}
\end{table}

Table~\ref{tab:appendix-prompt-schema} tests whether the descriptor values themselves are sufficient. The naive condition keeps the same LLM, instances, image, indexed history, output format, descriptor values and order, and block position. It only replaces the GF/LF evidence schema with a neutral descriptor serialization. Structured GF/LF evidence yields higher standard and affiliation F1 than naive serialization for both backbones. This controlled gap indicates that raw descriptor values alone do not account for the result in this setting. The finding supports the representation-level motivation: frequency information becomes more useful when the prompt distinguishes sequence-level periodic context from time-localized spectral evidence.

\subsection{Same-Feature Non-LLM Controls}
\label{app:nonllm-controls}
\begin{table}[t]
\centering
\scriptsize
\setlength{\tabcolsep}{3.0pt}
\resizebox{\columnwidth}{!}{%
\begin{tabular}{@{}lccrr@{}}
\toprule
Method & LLM & Same desc. & F1 & Affi. F1 \\
\midrule
FFT-ZScore & No & Yes & 34.63 & 75.19 \\
FFT-ZScore top 5\% & No & Yes & 21.28 & 52.61 \\
FFT-IsolationForest & No & Yes & 23.09 & 55.86 \\
FFT-CUSUM & No & Yes & 41.04 & 78.64 \\
LLM-TSAD, paired run & Yes & No & 67.14 & 82.57 \\
Ours LF+GF & Yes & Yes & \textbf{73.70} & \textbf{91.39} \\
\bottomrule
\end{tabular}%
}
\caption{Label-free non-LLM controls that use the same FFT-derived descriptors, preprocessing, and evaluation protocol. Thresholds are fixed a priori. The LLM-TSAD row is the reproduced paired diagnostic run used for these controls.}
\label{tab:appendix-nonllm-controls}
\end{table}

Table~\ref{tab:appendix-nonllm-controls} evaluates label-free classical controls that use the same FFT-derived descriptors and preprocessing without an LLM. Among these controls, FFT-CUSUM gives the highest standard F1 at 41.04 and affiliation F1 at 78.64, while LF+GF reaches 73.70 and 91.39 under the same evaluation protocol. These controls do not establish that an LLM is necessary for every frequency anomaly. They show more narrowly that the tested direct descriptor detectors do not reproduce the LF+GF result. This distinction is consistent with the method design, where FFT summaries provide auxiliary evidence and the LLM still performs interval prediction from the indexed sequence.

\subsection{Component-Level Descriptor Ablation}
\label{app:component-ablation}
\begin{table}[t]
\centering
\scriptsize
\setlength{\tabcolsep}{4pt}
\begin{tabular}{@{}lrr@{}}
\toprule
Added evidence & F1 & Affi. F1 \\
\midrule
LLM-TSAD baseline & 68.33 & 82.78 \\
Global dominant frequency & \textbf{74.01} & 89.45 \\
Global top-$k$ spectral peaks & 72.52 & 87.50 \\
Global spectral entropy & 71.91 & 86.61 \\
Global low/high energy ratio & 72.54 & 87.03 \\
Local dominant frequency & 72.70 & 87.91 \\
Local spectral entropy & 71.58 & 92.25 \\
Full LF+GF & 73.97 & \textbf{92.53} \\
\bottomrule
\end{tabular}
\caption{Component-level descriptor ablation on the balanced 1/4-scale diagnostic subset with Qwen2.5-VL-72B-Instruct. Each condition adds one descriptor type to the same baseline prompt; the final row uses the full structured LF+GF evidence.}
\label{tab:appendix-component-ablation}
\end{table}

Table~\ref{tab:appendix-component-ablation} isolates individual global and local descriptors on the balanced diagnostic subset. The descriptors show different metric profiles. Global dominant frequency gives the highest standard F1 among single descriptors, while local spectral entropy gives the highest affiliation F1 among single descriptors. No individual descriptor gives the strongest value on both metrics, and the full LF+GF configuration gives a balanced overall profile. This pattern matches the intended division of evidence: global descriptors summarize sequence-level periodic structure, whereas local descriptors expose interval-level variation. The result therefore supports complementary roles rather than a claim that every descriptor contributes uniformly.

\subsection{Evidence-Level Perturbation}
\label{app:evidence-perturbation}
\begin{table}[t]
\centering
\scriptsize
\setlength{\tabcolsep}{3pt}
\begin{tabularx}{\columnwidth}{@{}Xrrrr@{}}
\toprule
Condition & F1 & $\Delta$ F1 & Affi. F1 & $\Delta$ Affi. \\
\midrule
Correct GF/LF evidence & 73.70 & -- & 91.39 & -- \\
Local evidence shuffled & 71.50 & -2.20 & 87.84 & -3.55 \\
Global+local mismatched donor & 70.57 & -3.13 & 86.87 & -4.52 \\
Remove frequency evidence & 67.14 & -6.56 & 82.57 & -8.82 \\
\bottomrule
\end{tabularx}
\caption{Evidence-level perturbation with Qwen2.5-VL-72B-Instruct. The prompt remains matched except for the frequency-evidence blocks. The removal condition corresponds to the paired LLM-TSAD diagnostic run.}
\label{tab:appendix-evidence-perturbation}
\end{table}

Table~\ref{tab:appendix-evidence-perturbation} changes only the frequency-evidence blocks while keeping the remaining prompt matched. Shuffling local evidence reduces F1 from 73.70 to 71.50, using a mismatched global and local donor reduces it to 70.57, and removing frequency evidence reduces it to 67.14. Affiliation F1 follows the same ordering. The graded degradation is consistent with sensitivity to sequence-specific frequency information rather than a generic benefit from adding more text. This analysis remains an evidence-level intervention and does not identify the internal reasoning mechanism of the LLM.

\subsection{Local Window and Stride Sensitivity}
\label{app:window-sensitivity}
\begin{table}[t]
\centering
\scriptsize
\setlength{\tabcolsep}{7pt}
\begin{tabular}{@{}rrrr@{}}
\toprule
Window $L$ & Stride $S$ & F1 & Affi. F1 \\
\midrule
32 & 16 & 73.31 & 91.36 \\
64 & 32 & \textbf{73.70} & \textbf{91.39} \\
128 & 64 & 72.77 & 90.49 \\
\bottomrule
\end{tabular}
\caption{Full-scale sensitivity to local-window length with stride fixed to half of the window length.}
\label{tab:appendix-window-sensitivity}
\end{table}

Table~\ref{tab:appendix-window-sensitivity} evaluates local-window lengths of 32, 64, and 128 with stride fixed to half of the window length. Standard F1 ranges from 72.77 to 73.70, and affiliation F1 ranges from 90.49 to 91.39. The variation stays below one F1 point across this fourfold window range. This result suggests that the main result does not depend on a single narrowly selected temporal resolution.

\subsection{Type-Wise Reference and False-Positive Audit}
\label{app:typewise-audit}
\begin{table}[t]
\centering
\scriptsize
\setlength{\tabcolsep}{3pt}
\resizebox{\columnwidth}{!}{%
\begin{tabular}{@{}lrrrr@{}}
\toprule
Type & Indexed & LLM-TSAD (+D.S.) & Ours LF+GF & $\Delta$ vs. LLM-TSAD \\
\midrule
Frequency & 23.99 & 33.98 & 40.89 & +6.91 \\
Point & 88.13 & 77.51 & 83.30 & +5.79 \\
Range & 86.39 & 85.58 & 83.94 & -1.64 \\
Trend & 60.28 & 71.51 & 86.67 & +15.16 \\
Overall & 64.70 & 67.15 & 73.70 & +6.55 \\
\bottomrule
\end{tabular}%
}
\caption{Type-wise standard F1 with the de-seasonalized LLM-TSAD configuration as the direct reference for the frequency-evidence setting.}
\label{tab:appendix-typewise-direct}
\end{table}

For the frequency-evidence configuration, the direct reference is the de-seasonalized LLM-TSAD baseline because LF+GF augments that representation. Table~\ref{tab:appendix-typewise-direct} shows gains of +6.91 F1 for frequency anomalies, +5.79 for point anomalies, and +15.16 for trend anomalies, with a -1.64 change for range anomalies. The pattern supports a targeted interpretation rather than uniform dominance: frequency evidence contributes most when periodic, spectral, or long-range temporal structure is informative, while amplitude-defined categories can already contain strong time-domain cues.

\begin{table}[t]
\centering
\scriptsize
\setlength{\tabcolsep}{3pt}
\begin{tabularx}{\columnwidth}{@{}XXrr@{}}
\toprule
Analysis subset & Metric & LLM-TSAD & Ours LF+GF \\
\midrule
Fully normal sequences & Any-prediction false alarm rate & 0.20 & 2.17 \\
Non-frequency-dominant categories & Raw precision & 74.63 & 82.22 \\
Non-frequency-dominant categories & Spurious interval ratio & 3.85 & 2.81 \\
Non-frequency-dominant categories & Sequences with spurious interval & 7.28 & 5.86 \\
\bottomrule
\end{tabularx}
\caption{False-positive audit reported under the same evaluation protocol. Values follow the scale used by the evaluation audit.}
\label{tab:appendix-false-positive}
\end{table}

Table~\ref{tab:appendix-false-positive} checks whether the additional evidence simply induces broader anomaly prediction. On fully normal sequences, the any-prediction false alarm rate increases from 0.20 to 2.17. In the non-frequency-dominant categories, raw precision increases from 74.63 to 82.22, while both the spurious interval ratio and the number of sequences with a spurious interval decrease. The audit therefore shows a mixed but bounded effect: fully normal inputs become somewhat more sensitive, while the non-frequency-dominant anomaly categories do not show broad over-detection under these measures.

\subsection{Paired Bootstrap Confidence Intervals}
\label{app:bootstrap-ci}
\begin{table}[t]
\centering
\scriptsize
\setlength{\tabcolsep}{3pt}
\resizebox{\columnwidth}{!}{%
\begin{tabular}{@{}llrrl@{}}
\toprule
Backbone & Metric & LLM-TSAD & Ours & $\Delta$ 95\% CI \\
\midrule
Qwen2.5-VL-72B & Std. F1 & 67.14 & 73.70 & +6.56 [5.29, 7.72] \\
Qwen2.5-VL-72B & Affi. F1 & 82.57 & 91.39 & +8.82 [7.47, 10.09] \\
GPT-4o & Std. F1 & 69.03 & 70.20 & +1.17 [-0.19, 2.45] \\
GPT-4o & Affi. F1 & 79.98 & 87.35 & +7.37 [5.97, 8.69] \\
Gemini-2.5-Flash & Std. F1 & 76.62 & 77.97 & +1.35 [0.04, 2.60] \\
Gemini-2.5-Flash & Affi. F1 & 87.12 & 92.48 & +5.36 [4.28, 6.46] \\
InternVL2-Llama3-76B & Std. F1 & 38.85 & 42.15 & +3.30 [2.58, 4.03] \\
InternVL2-Llama3-76B & Affi. F1 & 69.93 & 72.41 & +2.47 [1.86, 3.13] \\
\bottomrule
\end{tabular}%
}
\caption{Paired bootstrap 95\% confidence intervals over evaluation instances. The LLM-TSAD values in this table are reproduced paired diagnostics for the confidence-interval analysis and do not replace the main-paper baseline values.}
\label{tab:appendix-bootstrap-ci}
\end{table}

Table~\ref{tab:appendix-bootstrap-ci} reports paired bootstrap 95\% confidence intervals over evaluation instances. Affiliation-F1 intervals remain above zero for all four backbones. Standard-F1 intervals remain above zero for Qwen2.5-VL-72B, Gemini-2.5-Flash, and InternVL2-Llama3-76B, while the GPT-4o interval includes zero. This result supports a more limited claim than uniform improvement: affiliation gains are stable in this paired analysis, whereas standard-F1 gains vary more by backbone. The baseline values in this table come from reproduced paired diagnostic runs and do not replace the main-paper baseline values.

\subsection{Newer Backbone Robustness Check}
\label{app:new-backbones}
\begin{table}[t]
\centering
\scriptsize
\setlength{\tabcolsep}{3pt}
\resizebox{\columnwidth}{!}{%
\begin{tabular}{@{}lrrrr@{}}
\toprule
Model & Baseline F1 / Affi. & Ours F1 / Affi. & $\Delta$ F1 & $\Delta$ Affi. \\
\midrule
GLM-4.6V & 51.58 / 75.10 & 57.78 / 76.87 & +6.20 & +1.77 \\
Gemini-3-Flash & 73.85 / 80.51 & 75.10 / 82.64 & +1.25 & +2.13 \\
GPT-5.1 & 75.77 / 85.64 & 76.80 / 88.79 & +1.03 & +3.15 \\
Average & -- & -- & +2.83 & +2.35 \\
\bottomrule
\end{tabular}%
}
\caption{Paired robustness check on three newer backbones using the same balanced 1/4-scale diagnostic subset. This check complements rather than replaces the full-scale main evaluation.}
\label{tab:appendix-new-backbones}
\end{table}

The main evaluation keeps the original backbone set to remain comparable with prior LLM-TSAD settings. Table~\ref{tab:appendix-new-backbones} adds a paired robustness check on GLM-4.6V, Gemini-3-Flash, and GPT-5.1 using the same balanced 1/4-scale subset. The average changes are +2.83 standard F1 and +2.35 affiliation F1. This diagnostic does not serve as a full-scale model leaderboard. It supports the narrower observation that the structured frequency-evidence design remains useful on the evaluated newer backbone set.

\subsection{Expanded TSB-AD-U Subset Evaluation}
\label{app:tsb-expanded}
\begin{table}[t]
\centering
\scriptsize
\setlength{\tabcolsep}{2.5pt}
\resizebox{\columnwidth}{!}{%
\begin{tabular}{@{}lrlrrrr@{}}
\toprule
Setting & Groups & Method & Std. & Affi. & $\Delta$ Std. & $\Delta$ Affi. \\
\midrule
Original & 8 & LLM-TSAD & 36.85 & 80.15 & -- & -- \\
Original & 8 & Ours & 40.26 & 80.35 & +3.41 & +0.20 \\
\midrule
Additional & 8 & LLM-TSAD & 13.51 & 68.46 & -- & -- \\
Additional & 8 & Ours & 16.14 & 70.00 & +2.63 & +1.54 \\
\midrule
Combined & 16 & LLM-TSAD & 25.18 & 74.31 & -- & -- \\
Combined & 16 & Ours & 28.20 & 75.18 & +3.02 & +0.87 \\
\bottomrule
\end{tabular}%
}
\caption{Expanded TSB-AD-U text+vision evaluation. The original eight-group subset follows the submission protocol, and the additional subset contains eight further groups. The combined row averages all 16 groups.}
\label{tab:appendix-tsb-expanded}
\end{table}

Table~\ref{tab:appendix-tsb-expanded} extends the original eight-group TSB-AD-U evaluation with eight additional groups under text+vision inference. The method improves the reported averages on both the original and additional subsets. Across all 16 groups, standard F1 changes from 25.18 to 28.20 and affiliation F1 changes from 74.31 to 75.18. The affiliation change is modest, so the result is best interpreted as a cross-subset robustness check rather than evidence of uniform gains on every TSB-AD-U group.

\section{Prompt Overhead and Evaluation Details}
\label{app:eval-details}

\subsection{Prompt-Length Audit}
\label{app:prompt-overhead}
\begin{table}[h!]
\centering
\scriptsize
\setlength{\tabcolsep}{3pt}
\resizebox{\columnwidth}{!}{%
\begin{tabular}{@{}lrrrrl@{}}
\toprule
Config. & Prompts & Avg. tokens & P95 & Max & Local windows / $\Delta$ \\
\midrule
LLM-TSAD & 1,600 & 8,137 & 8,137 & 8,137 & 0 / -- \\
Ours-LF & 1,600 & 8,854 & 8,854 & 8,854 & 30 / +717 (+8.8\%) \\
Ours-LF+GF & 1,600 & 8,970 & 8,971 & 8,973 & 30 / +834 (+10.2\%) \\
\bottomrule
\end{tabular}%
}
\caption{Prompt-length audit over all 4,800 prompts. Each configuration contains the same 1,000-point indexed history; LF and GF add the structured frequency-evidence text.}
\label{tab:appendix-prompt-overhead}
\end{table}

Table~\ref{tab:appendix-prompt-overhead} summarizes the token audit over 4,800 prompts. Because the baseline already contains the same 1,000-point indexed history, LF+GF increases the average text length from 8,137 to 8,970 tokens, corresponding to +834 tokens or +10.2\%. The LF-only condition adds +717 tokens or +8.8\%. These values quantify representation overhead rather than runtime overhead. Latency and provider-specific API cost can vary with serving conditions, so we do not infer them from token counts alone.

\subsection{Decoding, Parsing, and Metric Computation}
\label{app:eval-pipeline}
\begin{table}[h!]
\centering
\scriptsize
\setlength{\tabcolsep}{4pt}
\begin{tabularx}{\columnwidth}{@{}lX@{}}
\toprule
Item & Value \\
\midrule
Decoding & \texttt{temperature=0}, max output tokens = 512, request seed = 0 \\
Retry & At most 4 attempts; 30 s wait on HTTP 503; otherwise exponential backoff \\
Parsing & JSON array between the first \texttt{[} and the last \texttt{]}, with \texttt{start} and \texttt{end} fields \\
Invalid output & Retry; if all retries fail, assign all-zero metrics \\
Standard F1 & Point-wise scikit-learn precision, recall, and F1 \\
Affiliation F1 & \texttt{affiliation} package \\
Aggregation & Macro-average over instances, then categories \\
\bottomrule
\end{tabularx}
\caption{Evaluation and response-processing details used for the reproducibility analysis.}
\label{tab:appendix-eval-pipeline}
\end{table}

Table~\ref{tab:appendix-eval-pipeline} specifies the response-processing and evaluation pipeline used in the reproducibility analysis. Decoding uses temperature 0, a maximum of 512 output tokens, and request seed 0. A request receives at most four attempts. HTTP 503 responses use a 30-second wait, while other retryable failures use exponential backoff. Parsing extracts the JSON array between the first opening bracket and the last closing bracket and requires \texttt{start} and \texttt{end} fields. If all retries fail, the instance receives all-zero metrics. Standard precision, recall, and F1 use point-wise scikit-learn computation, affiliation metrics use the \texttt{affiliation} package, and aggregation takes a macro-average over instances and then categories.

\end{document}